\documentclass[11pt]{article}

\usepackage[preprint]{acl}

\usepackage{times}
\usepackage{latexsym}
\usepackage[T1]{fontenc}
\usepackage[utf8]{inputenc}
\usepackage{microtype}
\usepackage{inconsolata}
\usepackage{graphicx}

\usepackage{amsmath}
\usepackage{booktabs}
\usepackage{algorithm}
\usepackage{algorithmic}
\usepackage{listings}
\usepackage{stfloats}
\usepackage{enumitem}

\lstdefinestyle{promptbox}{
  basicstyle={\small\ttfamily},
  numbers=none,
  frame=single,
  framerule=0.5pt,
  framesep=7pt,
  xleftmargin=0pt,
  framexleftmargin=0pt,
  framexrightmargin=0pt,
  showstringspaces=false,
  columns=fullflexible,
  keepspaces=true,
  tabsize=2,
  breaklines=true,
  breakatwhitespace=false,
  aboveskip=7pt,
  belowskip=7pt,
  captionpos=b
}

\title{MDB-Link: Hierarchical Schema Linking for Multi-Database Text-to-SQL}

\author{Beiyu Xu\and Zhenyu Wu \and Jiaoyan Chen \and Riza theresa Batista-navarro\\
University of Manchester \\
  Faculty of Science and Engineering \\
  Department of Computer Science \\}

\begin{document}

\maketitle

\begin{abstract}
Traditional Text-to-SQL research and benchmarks assume a known target database, overlooking settings in which a query must be routed within a large, heterogeneous database collection. We therefore study schema linking in a multi-database setting, where the system must first locate the target database and then construct a compact, SQL-relevant schema for generation. We propose MDB-Link, a hierarchical schema-linking framework that retrieves question-relevant columns from a global index, aggregates retrieval evidence to shortlist databases, and uses a budget-aware large language model (LLM) for database reranking, table selection, and column grounding. 
With Qwen2.5-14B, MDB-Link outperforms LinkAlign on MMQA, Spider2-Snow, and BIRD-dev in database localization and column selection while producing schema subsets close in size to the gold schemas. 
Exact match improves from 16.88 to 51.41 on MMQA, 2.50 to 9.17 on Spider2-Snow, and 12.52 to 38.01 on BIRD-dev. MDB-Link also runs faster than LinkAlign and AutoLink, demonstrating the effectiveness of hierarchical schema reduction for downstream SQL generation.

\end{abstract}

\section{Introduction}
Traditional Text-to-SQL research and benchmarks generally assume that each question is paired with its target database, confining SQL generation to a known schema~\cite{wang2020rat,li2023bird}. Although this setup supports controlled evaluation, it omits database localization: in enterprise data platforms, the target database may be unspecified and must be identified from a large, heterogeneous collection at inference time \cite{lei2025spider,wu2025mmqa}. In this setting, a system must determine not only how to express a question in SQL, but also where the required data reside and which schema elements are needed. We refer to this upstream task as \emph{multi-database schema linking}~\cite{wang2025linkalign}.
This task has a hierarchical structure: a system must first localize a database from a large collection, then identify the relevant tables, and finally ground the columns needed for SQL generation. These decisions are interdependent: choosing the wrong database invalidates all downstream selections, while omitting a bridge table or join key can prevent correct SQL generation even when other selected columns are semantically related. 
The central challenge is therefore to narrow the candidates at each level without removing schema items required by the correct query or exceeding the context budget of the downstream SQL generator.

Existing strategies expose complementary limitations. Full-schema prompting can be effective for small schemas, but its context cost and exposure to irrelevant schema elements grow with the number and size of candidate databases \cite{pourreza2023din,sun2023sqlprompt,gao2023text}. Retrieval and ranking methods reduce this burden, yet flat item scoring may overlook structurally necessary tables or columns that are only weakly expressed in the question \cite{wang2020rat,li2023resdsql,katsogiannis2023survey}. Agentic exploration can recover additional schema evidence, but broad expansion may increase latency and schema noise \cite{wang2025linkalign,wang2026autolink}. These trade-offs motivate a method that combines explicit database localization with structured, budget-aware schema reduction.

To address this need, we propose \textbf{MDB-Link}, a hierarchical framework that uses question-relevant column evidence from a global index to localize candidate databases and then narrows the predicted database through table selection and column grounding. A budget-aware schema context construction strategy preserves relational metadata while limiting the schema context passed to the LLM. By separating these decisions, MDB-Link aims to avoid both exhaustive full-schema prompting and unconstrained schema expansion.

Our contributions are summarized as follows:
\begin{itemize}
    \item We formulate multi-database schema linking as a hierarchical grounding problem that requires progressive database-, table-, and column-wise schema-space reduction before SQL generation.
    \item We propose \textbf{MDB-Link}, a hierarchical multi-database schema linking framework that integrates database localization based on global column evidence retrieval, LLM logits-based database reranking, table selection, column-wise grounding, and budget-aware schema context construction.
    \item We introduce an evaluation protocol that jointly considers target-database accuracy, column exact match and recall, schema compactness, inference cost, and downstream SQL utility. Experiments on MMQA, Spider2-Snow, and BIRD-dev show that MDB-Link offers a favorable balance across these dimensions compared with the evaluated baselines.
\end{itemize}

\section{Related Work}
Prior schema-linking research mainly considers settings in which the target database schema is already provided. Encoder-based parsers, including RAT-SQL~\cite{wang2020rat}, LGESQL~\cite{cao2021lgesql}, and S2SQL~\cite{hui2022s2sql}, model question--schema relations through relational or graph-based representations. LLM-based systems such as RESDSQL~\cite{li2023resdsql}, DIN-SQL~\cite{pourreza2023din}, and CHESS~\cite{talaei2024chess} instead use selectors, rankers, decomposition, or metadata augmentation to construct a schema context before SQL generation. These methods improve table- and column-level grounding within a known database, but they do not select the target database from a larger collection.

Research on large-schema Text-to-SQL has introduced full-schema prompting, retrieval, reranking, graph search, and agentic exploration. Full-schema prompting delegates grounding to the SQL generator, but its context cost grows with schema size \cite{sun2023sqlprompt,gao2023text}. Retrieval and graph-based methods reduce this burden by selecting candidate schema items, as in SchemaGraphSQL~\cite{safdarian26schemagraphsql}. LinkAlign~\cite{wang2025linkalign} combines database retrieval with schema-item grounding for multi-database settings, whereas AutoLink~\cite{wang2026autolink} iteratively explores large, pre-selected schemas; ReFoRCE~\cite{deng2025reforce} similarly uses iterative column exploration during SQL generation. These approaches improve schema coverage, but their different expansion strategies leave open how to jointly control database localization, structural completeness, schema compactness, and inference cost. MDB-Link addresses this gap by making database localization an explicit first stage and then constructing a budgeted table--column context within the predicted database.

\section{Problem Statement}
We study schema linking in a multi-database Text-to-SQL setting. Given a natural language question \(q\), an optional hint or external knowledge \(h\), and a collection of databases \(\mathcal{D}=\{D_1,D_2,\ldots,D_n\}\), the goal is to identify the database that contains the information needed to answer \(q\), and then select the schema items within that database that are necessary for SQL generation. Each database \(D_i\) contains a set of tables, columns, and available schema metadata, including column descriptions, sample values, value descriptions, primary keys, and foreign-key relationships.

Formally, the schema linker produces a target database prediction \(\hat{D}\in\mathcal{D}\) and a linked schema subset \(\hat{S}\). The linked schema subset is organized by tables and columns:
\[
\hat{S}=\{(t, C_t)\mid t \in T_{\hat{D}}, C_t \subseteq C(t)\},
\]
where \(T_{\hat{D}}\) denotes the tables in the predicted database and \(C(t)\) denotes the columns of table \(t\). The output should contain the target database, its tables, and columns required for generating a SQL query that can correctly answer the question $q$, including predicates, projections, aggregation, ordering, and joins, while excluding irrelevant schema items as much as possible.

\begin{figure*}[ht]
\centering
\includegraphics[width=0.98\textwidth]{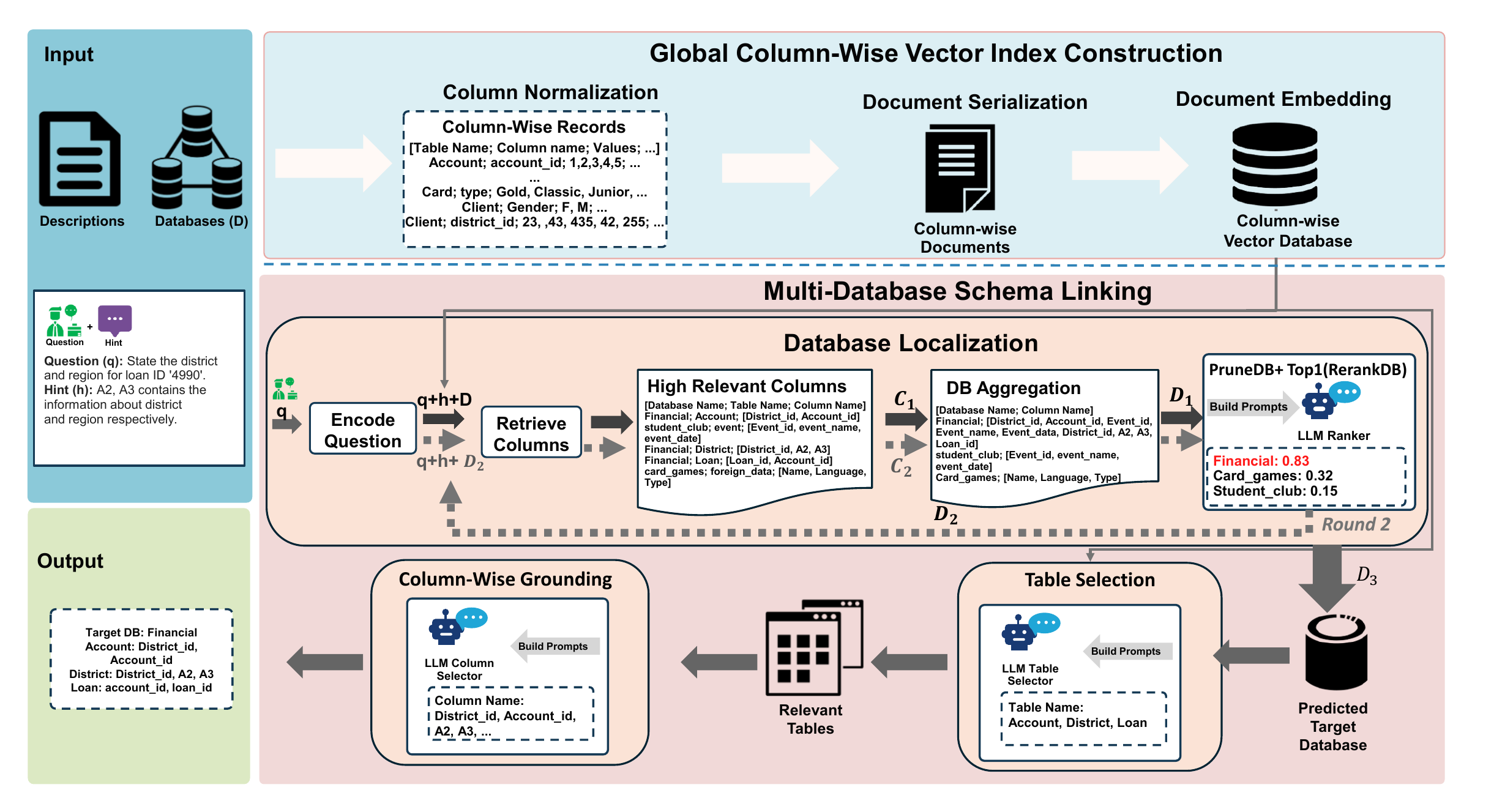}
\caption{The MDB-Link framework comprises offline construction of a global column-wise vector index and online multi-database schema linking through database localization, table selection, and column-wise grounding.}
\label{workflow}
\end{figure*}

\section{Methodology} 
As shown in Figure~\ref{workflow}, MDB-Link comprises two stages: offline vector index construction and online hierarchical schema linking. Offline, it normalizes and serializes column-wise records from all candidate databases, embeds them, and stores their vectors and metadata in a persistent local vector database with a searchable index. Online, MDB-Link proceeds through three modules: database localization, table selection, and column-wise grounding. 
Database localization uses retrieved column evidence to identify and rerank candidate databases, then selects relevant tables in the predicted database, and finally grounds the columns required for SQL generation. 
The LLM-based modules (database reranking, table selection, and column-wise grounding) follow a common budget-aware schema context construction strategy. Database reranking and table selection use a shared full-schema-first renderer, which first tests a prompt containing every table and column of the database being processed. Column-wise grounding uses a selected-table context procedure, which first attempts to use a prompt containing every column of the selected tables. Across all modules, the schema context is reduced only if the resulting prompt exceeds the token budget.

\subsection{Global Column-Wise Vector Index Construction}
For each database collection \(\mathcal D\), corresponding to one dataset in our experiments, MDB-Link builds a single global vector index over all columns from its candidate databases. 
Following dense retrieval and retrieval-augmented generation, the index supports collection-wide semantic matching between a question and column-level text  \cite{karpukhin2020dense,lewis2020retrieval}. 
Columns are used as retrieval units because they are the finest-grained schema elements selected downstream and permit direct alignment with question semantics. Moreover, available column and value descriptions supplement table and column names, while database and table provenance allows each retrieved column to be mapped back to its parent schema.

Index construction has three steps. 
First, MDB-Link creates one normalized record per column, storing its database and table provenance; column name, type, description, sample values, and value descriptions; and available primary- and foreign-key metadata. 
Second, it serializes the table name, column name, and available column and value descriptions into a compact retrieval document. Only this document is embedded; the complete normalized record is retained as structured payload. 
Third, each retrieval document is encoded using an embedding model, and the resulting vector and payload are stored in a persistent local Qdrant collection configured with cosine distance \cite{qdrant2026}.
The serialized retrieval documents and selected payload fields for \texttt{Account.Date} and \texttt{Card.Type} in Figure~\ref{workflow} are provided in Appendix~\ref{app:embedding_text_example}.

\subsection{Database Localization}
This module identifies the database whose schema contains the information needed to answer the input question. As summarized in Algorithm~\ref{alg:database_localisation}, MDB-Link uses an adaptive two-round strategy: it first retrieves column evidence across the complete database collection to reduce the risk of missing the target, and then progressively narrows the candidate set so that LLM reranking is applied only to the retained databases.

MDB-Link first encodes the question \(q\) once to obtain the query vector \(\mathbf e_q\) (Line~\ref{line:encodeq}), which is reused across two rounds. 
In the first round, it searches the entire database collection \(\mathcal D\): computing the retrieval budget \(b_1\), retrieving the top-\(b_1\) highly relevant columns (HRC) \(\mathcal C_1\), and grouping this column evidence by database before support-based pruning produces the candidate set \(\mathcal D_1\) (Lines~\ref{line:compB}--\ref{line:PruneDBD1}). If \(|\mathcal D_1|\leq\kappa\), \(\mathrm{RerankDB}\) ranks \(\mathcal D_1\), and its top-ranked database is selected as the target database \(\hat{D}\) and returned (Lines~\ref{line:if}--\ref{line:RerankDB1}). Otherwise, the algorithm takes the second round for further database pruning.
RerankDB retains the top-\(\kappa\) databases as \(\mathcal D_2\), which becomes the search scope for the second round (Lines~\ref{line:else}--\ref{line:RerankDBk}). It reuses \(\mathbf e_q\), computes \(b_2\) over \(\mathcal D_2\), retrieves \(\mathcal C_2\) only from \(\mathcal D_2\), and applies the same support-based pruning to produce \(\mathcal D_3\subseteq\mathcal D_2\) (Lines~\ref{line:ComputeBudget2}--\ref{line:PruneD3}). Finally, \(\mathrm{RerankDB}\) ranks \(\mathcal D_3\), and the top-ranked database is returned as the target database \(\hat{D}\) (Lines~\ref{line:RerankDBf}--\ref{line:end}).
\begin{algorithm}[ht]
\small
\caption{Database Localization in MDB-Link}
\label{alg:database_localisation}
\begin{algorithmic}[1]
\REQUIRE Question \(q\), optional hint \(h\), databases \(\mathcal D\), column index \(\mathcal I\), second-round pruning threshold \(\kappa\)
\ENSURE Predicted target database \(\hat{D}\)
\STATE \(\mathbf e_q \leftarrow \mathrm{EncodeQuestion}(q)\)\label{line:encodeq}
\STATE \(b_1 \leftarrow \mathrm{ComputeBudget}( \mathcal D )\)\label{line:compB}
\STATE \(\mathcal C_1 \leftarrow \mathrm{RetrieveColumns}(\mathcal I,\mathbf e_q,\mathcal D)\)\label{line:retrCols}
\STATE \(\mathcal D_1 \leftarrow \mathrm{PruneDB}(\mathcal C_1)\)\label{line:PruneDBD1}
\IF{\(|\mathcal D_1| \leq \kappa\)}\label{line:if}
    \STATE \(\hat{D} \leftarrow \mathrm{Top}_{1}(\mathrm{RerankDB}(\mathcal D_1,q,h))\)\label{line:RerankDB1}
\ELSE\label{line:else}
    \STATE \(\mathcal D_2 \leftarrow \mathrm{Top}_{\kappa}(\mathrm{RerankDB}(\mathcal D_1,q,h))\)\label{line:RerankDBk}
    \STATE \(b_2 \leftarrow \mathrm{ComputeBudget}(\mathcal D_2)\)\label{line:ComputeBudget2}
    \STATE \(\mathcal C_2 \leftarrow \mathrm{RetrieveColumns}(\mathcal I,\mathbf e_q,\mathcal D_2)\)\label{line:RetrieveColumns2}
    \STATE \(\mathcal D_3 \leftarrow \mathrm{PruneDB}(\mathcal C_2)\)\label{line:PruneD3}
    \STATE \(\hat{D} \leftarrow \mathrm{Top}_{1}(\mathrm{RerankDB}(\mathcal D_3,q,h))\)\label{line:RerankDBf}
\ENDIF
\RETURN \(\hat{D}\)\label{line:end}
\end{algorithmic}
\end{algorithm}

\(\mathrm{EncodeQuestion}(q)\) takes the question \(q\) as input and applies the same embedding model used to construct \(\mathcal I\), producing the query vector \(\mathbf e_q\).

The algorithm uses a budget-aware column retrieval strategy which first uses \(\mathrm{ComputeBudget}\) to calculate budgets (i.e., the maximum number of HRCs) and then uses \(\mathrm{RetrieveColumns}\) to retrieve columns over a collection of databases.
For a set of databases \(D\), which has in total $N$ columns,
 \(\mathrm{ComputeBudget}\) returns \( b_i=\min\!\left(\beta_i,\left\lceil\alpha N\right\rceil\right)\)
, where \(\alpha\) and \(\beta_i\) denote the ratio of $N$ and maximum number of the column to return, respectively.
The function \(\mathrm{RetrieveColumns}(\mathcal I,\mathbf e_q, D,b_i)\) then performs an exact cosine-similarity search over \(\mathcal I\): it compares \(\mathbf e_q\) with every indexed column vector whose stored database identifier belongs to \(\mathcal D\), and returns the top-\(b_i\) scored columns as \(\mathcal C_i\).

\(\mathrm{PruneDB}\) takes a set of scored columns \(\mathcal C\) and returns an ordered list of databases. Suppose that the columns in \(\mathcal C\) come from \(M\) distinct databases. For each database \(D_i\), it computes the hit count \(n_i\), maximum similarity \(m_i\), and score sum \(a_i\). It then ranks the databases using lexicographic ordering: \(m_i\), \(a_i\), and \(n_i\) are compared in descending order, while the database identifier is compared in ascending lexicographic order as the final tie-breaker. The ordered list retains \(D_i\) if
\[
n_i \geq \eta
\quad\mathrm{or}\quad
m_i \geq Q_{\rho}(m_1,\ldots,m_M),
\]
where \(Q_{\rho}\) denotes the \(\rho\)-quantile operator over the database-level maximum similarities. After support filtering, the function returns at most the first \(\mu\) databases in this order. This procedure removes weak long-tail matches while preserving databases supported by either repeated hits or one highly confident column.

\(\mathrm{RerankDB}\) returns an ordered list of databases based on the score by the LLM reranker. For each candidate database, it constructs a binary reranking prompt using the shared full-schema-first renderer described in Section~\ref{sec:budget_context}. The prompt asks whether the database schema contains sufficient information to answer the question and restricts the output to \texttt{yes} or \texttt{no}; the prompt template and an example candidate-database schema context are provided in Appendices~\ref{app:database_reranking_prompt} and~\ref{app:database_reranking_example}. Following prior work on using constrained token probabilities as confidence estimates \cite{kadavath2022language,si2022prompting}, MDB-Link scores each database by the normalized probability of the positive label. Let \(\mathcal{V}_{\texttt{yes}}\) and \(\mathcal{V}_{\texttt{no}}\) denote the sets of token IDs that individually decode, after whitespace stripping and lowercasing, to \texttt{yes} and \texttt{no}, respectively. Given next-token logits \(z_v\), the database relevance score is
\[
\resizebox{\columnwidth}{!}{$\displaystyle
s(D_i\mid q,h)=
\frac{\sum_{v\in \mathcal{V}_{\texttt{yes}}}\exp(z_v)}
{\sum_{v\in \mathcal{V}_{\texttt{yes}}}\exp(z_v)+
 \sum_{v\in \mathcal{V}_{\texttt{no}}}\exp(z_v)}.
$}
\]
The function returns the candidate databases ranked by \(s(D_i\mid q,h)\), after which \(\mathrm{Top}_1\) or \(\mathrm{Top}_{\kappa}\) selects the number required by the corresponding branch of Algorithm~\ref{alg:database_localisation}.

\subsection{Table Selection}
Given the predicted database \(\hat{D}\), question \(q\), and optional hint \(h\), table selection predicts a valid set of tables predicted as relevant\(\hat{\mathcal{T}}\subseteq T_{\hat{D}}\). When \(\hat{\mathcal{T}}\) is nonempty, column-wise grounding searches only within these tables. This step is needed because database localization identifies only the target database, which may still contain many irrelevant tables. Excluding such tables from column-wise grounding reduces schema noise and narrows the subsequent search space.

MDB-Link constructs the table-selection prompt using the shared full-schema-first renderer described in Section~\ref{sec:budget_context}. The LLM predicts the relevant table names. These names are deduplicated while preserving their first-occurrence order and validated by exact matching against the tables in \(\hat{D}\); the surviving names form \(\hat{\mathcal{T}}\). If no valid table remains, column-wise grounding falls back to the schema context retained during table selection rather than receiving an empty context. The prompt template is shown in Appendix~\ref{app:table_selection_prompt}.

\subsection{Column-Wise Grounding}
After the relevant tables are selected, MDB-Link performs column-wise grounding within the reduced table scope. This stage identifies the specific columns required to construct the SQL query, including columns used for projection, filtering, aggregation, ordering, and joins. 

MDB-Link constructs the column-grounding prompt over \(\hat{\mathcal{T}}\) using the selected-table context procedure described in Section~\ref{sec:budget_context}. The LLM returns the necessary columns as a table-to-column mapping, with all names constrained to the displayed schema context and join keys included only when needed. The prompt template is shown in Appendix~\ref{app:column_grounding_prompt}.

The output of this stage is the final linked schema subset. It contains the predicted target database, the selected tables, and the grounded columns organized by table. This linked schema is then passed to the downstream SQL generation model as a compact representation of the schema evidence needed to answer the question.

\subsection{Budget-Aware Schema Context Construction}
\label{sec:budget_context}
Database reranking, table selection, and column-wise grounding follow a common budget-aware schema context construction strategy. In all three steps, MDB-Link applies the token budget to the complete prompt, including the prompt template, question, optional hint, and schema context, rather than to the schema context alone.

Database reranking and table selection use the shared full-schema-first renderer. It first attempts to include every table and column of the current database. If the prompt exceeds the input-token budget, it ranks columns by their cosine similarity to \(q\) using the global column-wise vector index, retains the highest-ranked column from each table, prioritizes primary- and foreign-key columns, and adds further columns across tables within the budget. Column-wise grounding uses the selected-table context procedure: it first attempts to include every column of the selected tables and otherwise reuses the table-selection records retained for those tables when available. Detailed selection and fallback procedures are provided in Appendix~\ref{app:budget_aware_context_procedure} and Algorithms~\ref{alg:budget_aware_context} and~\ref{alg:column_context_fallback}.

\section{Experiments}
\subsection{Datasets}
We evaluate MDB-Link on three multi-database Text-to-SQL benchmarks: MMQA \cite{wu2025mmqa}, Spider2-Snow, which is a Snowflake-oriented subset with gold SQL queries derived from Spider 2.0 \cite{lei2025spider}, and the development split of BIRD \cite{li2023bird}. For MMQA, we recover the source SQLite databases from Spider \cite{yu2018spider} and exclude instances with invalid gold answers. Table~\ref{tab:dataset_stats} reports the numbers of samples (questions), databases, tables, and columns in each benchmark. More detailed schema statistics are provided in Appendix~\ref{app:detailed_dataset_statistics}.

\begin{table}[ht]
\centering
\small
\begin{tabular}{lrrrr}
\toprule
Dataset & Samples & DBs & Tables & Columns \\
\midrule
MMQA & 3,128 & 158 & 808 & 4,220 \\
Spider2-Snow & 120 & 53 & 1,938 & 102,341 \\
BIRD (dev) & 1,534 & 11 & 75 & 798 \\
\bottomrule
\end{tabular}
\caption{Statistics of the three evaluation datasets.}
\label{tab:dataset_stats}
\end{table}

\subsection{Baselines}
We compare MDB-Link with four baselines:
\begin{itemize}[leftmargin=*]
    \item \textbf{LinkAlign} \cite{wang2025linkalign} is a representative and scalable multi-database schema-linking baseline that adopts iterative alignment and filtering.
    \item \textbf{AutoLink} \cite{wang2026autolink} is a state-of-the-art schema-linking method that expands candidate schemas through iterative exploration and an agentic strategy.
    \item \textbf{Single-Prompt} uses direct LLM prompting without MDB-Link's hierarchical reduction; its complete prompts are provided in Appendix~\ref{app:single_prompt_prompt}.
    \item \textbf{Dense Retrieval} is an embedding-based retrieval baseline. It encodes the question and the columns, and ranks the columns by cosine similarity. Following the initial Number of Columns to Retrieve setting in Autolink, the top 20 columns are used as the schema-linking result. For database localization, the predicted database is the database that receives the largest number of hits among the top 20 retrieved columns.
\end{itemize}

\subsection{Evaluation Metrics}
We evaluate MDB-Link at both the schema-linking and downstream SQL-generation levels using the following metrics:
\begin{itemize}[leftmargin=*]
    \item \textbf{Locate Accuracy (LA)} is the proportion of samples for which the predicted database exactly matches the gold target database.
    \item \textbf{Exact Match (EM)} measures the proportion of samples whose predicted schema is the same as the gold schema.
    \item \textbf{Recall} is computed at the micro level over all samples and is the fraction of gold columns covered by the predicted columns.
    \item \textbf{Avg. Predicted Columns (\#Cols)}  measures schema compactness and is the average number of columns selected by the linker.
    \item \textbf{Avg Token (Tok.) and Avg Time (Time)} measure schema-linking efficiency. Avg Token is the average total number of input and output LLM tokens consumed per sample by the entire database-localization and schema-linking method. Avg Time is the average online end-to-end runtime per sample, including online retrieval and all LLM inference calls; the one-time offline vector-index construction is excluded.
    \item \textbf{Execution Accuracy (EX)} measures downstream correctness by comparing the execution result of the predicted SQL with that of the gold SQL on the corresponding database. \textbf{Avg. Tokens} and \textbf{Avg. Time} measure downstream efficiency as the average Spider-Agent token consumption and runtime per sample, respectively.
\end{itemize}

\subsection{Implementation Details}
We use Qwen3-Embedding-0.6B \cite{qwen3embedding} to construct the column-wise vector index and compute embedding-based retrieval scores. For LLM-based database reranking, table selection, column-wise grounding, and LLM-based baselines, we evaluate Qwen2.5-14B-Instruct \cite{qwen2.5} and Ministral-3-14B-Instruct \cite{liu2026ministral}. For downstream Text-to-SQL evaluation, we use Spider-Agent from Spider 2.0 \cite{lei2025spider} with Qwen3-Coder-30B-A3B-Instruct \cite{qwen3technicalreport} as its SQL-generation backbone. All fixed hyperparameter and inference settings are provided in Appendix~\ref{app:implementation_settings}. All experiments are conducted on a single NVIDIA A100 GPU with 80 GB memory.

\section{Main Results}

\begin{table*}[t]
\centering
\small
\setlength{\tabcolsep}{3.5pt}
\begin{tabular*}{\textwidth}{@{\extracolsep{\fill}}lrrrrrrrrrrrr@{}}
\toprule
& \multicolumn{4}{c}{Spider2-Snow} & \multicolumn{4}{c}{MMQA} & \multicolumn{4}{c}{BIRD-dev} \\
\cmidrule(lr){2-5}\cmidrule(lr){6-9}\cmidrule(lr){10-13}
Method & LA & EM & Recall & \#Cols & LA & EM & Recall & \#Cols & LA & EM & Recall & \#Cols \\
\midrule
LinkAlign (M3) & 51.67 & 5.00 & 33.56 & 10.60 & 87.40 & 34.24 & 85.09 & 6.50 & 98.15 & 22.62 & 85.94 & 6.44 \\
LinkAlign (Q2.5) & 50.00 & 2.50 & 20.07 & 5.78 & 88.68 & 16.88 & 66.75 & 4.79 & 95.44 & 12.52 & 70.42 & 4.72 \\
AutoLink (M3) & -- & -- & 65.43 & 197.86 & -- & -- & \textbf{96.67} & 72.65 & -- & -- & 97.35 & 44.43 \\
AutoLink (Q2.5) & -- & -- & \textbf{66.27} & 235.98 & -- & -- & 95.89 & 89.40 & -- & -- & \textbf{97.49} & 52.90 \\
Single-Prompt (M3) & -- & -- & -- & -- & 74.04 & 30.53 & 68.69 & 5.10 & 95.50 & 32.14 & 82.28 & 4.77 \\
Single-Prompt (Q2.5) & -- & -- & -- & -- & 59.85 & 28.61 & 55.50 & 4.74 & 98.24 & 36.77 & 80.97 & 4.48 \\
Dense Retrieval (Q3E) & 36.67 & 0.00 & 14.17 & 20.00 & 40.92 & 0.00 & 38.18 & 19.44 & 88.01 & 0.00 & 57.62 & 19.23 \\
\midrule
\textbf{MDB-Link (M3)} & 60.00 & 1.67 & 39.12 & 10.16 & \textbf{90.54} & 43.00 & 83.22 & 5.34 & \textbf{98.89} & \textbf{39.05} & 82.72 & 4.69 \\
\textbf{MDB-Link (Q2.5)} & \textbf{65.00} & \textbf{9.17} & 32.10 & 9.49 & 89.71 & \textbf{51.41} & 86.32 & 5.31 & \textbf{98.89} & 38.01 & 81.19 & 4.38 \\
\bottomrule
\end{tabular*}
\caption{Overall schema linking performance on Spider2-Snow, MMQA, and BIRD-dev. For LLM-based methods, M3 and Q2.5 denote the 14B versions of Ministral-3-Instruct and Qwen2.5-Instruct, respectively; Q3E denotes Qwen3-Embedding-0.6B. LA, EM, and Recall are reported as percentages. The average numbers of gold columns are 9.88, 5.28, and 4.47 for Spider2-Snow, MMQA, and BIRD-dev, respectively. For Spider2, the Single-Prompt cannot be conducted because the prompt is out-of-window.}
\label{tab:overall_schema_linking}
\begin{minipage}{\textwidth}
\footnotesize\emph{Note.} AutoLink may return columns from multiple databases and therefore does not produce a single target-database prediction, so LA is not applicable. EM is not reported for AutoLink because its multi-database schema output is not evaluated as a single-database linked column set in our protocol.
\end{minipage}
\end{table*}

\subsection{Overall Schema Linking Performance}
Table~\ref{tab:overall_schema_linking} summarizes overall schema-linking performance. Against LinkAlign under the same backbone, MDB-Link improves LA in every comparison. EM also improves except on Spider2-Snow with Ministral-3-14B, where it decreases from 5.00 to 1.67 even as LA and recall increase. Recall is higher in every comparison except for MMQA and BIRD-dev with Ministral-3-14B, where the decreases are modest. The largest EM gains occur with Qwen2.5-14B: EM rises from 16.88 to 51.41 on MMQA and from 2.50 to 9.17 on Spider2-Snow, while MDB-Link reaches 98.89 LA on BIRD-dev under both backbones.

AutoLink obtains the highest recall but returns 9.5--24.9$\times$ more columns than MDB-Link. The predicted schema from AutoLink includes a large ratio of irrelevant columns, which requires a more powerful downstream generation method. Single-Prompt remains feasible on MMQA and BIRD-dev but cannot process Spider2-Snow, while Dense Retrieval yields low LA and recall on all three datasets. Overall, MDB-Link provides the most consistent improvements in database localization and exact schema grounding while keeping the linked schema compact. The results therefore support a stronger accuracy--compactness trade-off, rather than uniform superiority on every metric.

\subsection{Efficiency Analysis}
Table~\ref{tab:efficiency_comparison} reports average token usage and online runtime. Complete efficiency results for both Ministral-3-14B and Qwen2.5-14B are provided in Appendix~\ref{app:complete_efficiency_results}. Because token demand depends on both schema scale and method-specific interaction patterns, we compare methods only within the same dataset and backbone. We interpret token usage jointly with online runtime and schema-linking quality (i.e., results in Table~\ref{tab:overall_schema_linking}).
\begin{table}[t]
\centering
\small
\setlength{\tabcolsep}{1.2pt}
\begin{tabular*}{\columnwidth}{@{\extracolsep{\fill}}lrrrrrr@{}}
\toprule
& \multicolumn{2}{c}{Spider2-Snow} & \multicolumn{2}{c}{MMQA} & \multicolumn{2}{c}{BIRD-dev} \\
\cmidrule(lr){2-3}\cmidrule(lr){4-5}\cmidrule(lr){6-7}
Method & Tok. & Time & Tok. & Time & Tok. & Time \\
\midrule
LinkAlign & \textbf{223} & 409.72 & 118 & 119.35 & 122 & 127.62 \\
AutoLink & 652 & 465.42 & 118 & 64.84 & 48 & 45.22 \\
Single-Prompt & -- & -- & 45 & \textbf{12.28} & \textbf{13} & \textbf{3.82} \\
\midrule
\textbf{MDB-Link} & 515 & \textbf{369.17} & \textbf{40} & 14.15 & 58 & 12.30 \\
\bottomrule
\end{tabular*}
\caption{Efficiency comparison under Qwen2.5-14B. Tokens are reported in thousands, and Time is reported in seconds.}
\label{tab:efficiency_comparison}
\end{table}
Under Qwen2.5-14B, MDB-Link is faster than LinkAlign and AutoLink on every dataset. Compared with LinkAlign on Spider2-Snow, MDB-Link uses more tokens but reduces the average runtime from 409.72 to 369.17 seconds while improving LA, EM, and Recall. On MMQA and BIRD-dev, it uses fewer tokens, runs faster, and improves all three schema-linking metrics. These results show that token usage and runtime capture different aspects of online efficiency. Compared with AutoLink, MDB-Link runs faster and returns far fewer columns on all three datasets; it uses fewer tokens on Spider2-Snow and MMQA but more on BIRD-dev, whereas AutoLink attains higher Recall with substantially broader schemas. Single-Prompt has the lowest runtime where feasible, but yields lower LA, EM, and Recall and cannot process Spider2-Snow. Overall, MDB-Link provides a favorable balance between online efficiency and schema-linking quality rather than uniformly minimizing every efficiency measure.

\begin{table*}[t]
\centering
\small
\setlength{\tabcolsep}{4.2pt}
\begin{tabular*}{\textwidth}{@{\extracolsep{\fill}}lrrrrrrrrr@{}}
\toprule
& \multicolumn{3}{c}{Spider2-Snow} & \multicolumn{3}{c}{MMQA} & \multicolumn{3}{c}{BIRD-dev} \\
\cmidrule(lr){2-4}\cmidrule(lr){5-7}\cmidrule(lr){8-10}
Method & EX & Tok. & Time & EX & Tok. & Time & EX & Tok. & Time \\
\midrule
LinkAlign & 7.50 & \textbf{46,235} & 181.56 & 55.40 & 7,169 & 30.58 & 50.00 & 7,248 & 26.47 \\
AutoLink & 7.50 & 103,307 & 163.50 & \textbf{69.18} & 7,662 & \textbf{22.79} & \textbf{57.69} & 12,824 & 25.71 \\
\midrule
\textbf{MDB-Link} & \textbf{8.30} & 55,532 & \textbf{144.58} & 63.04 & \textbf{5,567} & 25.71 & 55.35 & \textbf{6,077} & \textbf{21.53} \\
\bottomrule
\end{tabular*}
\caption{Downstream Spider-Agent results with Qwen2.5-14B schema links. Avg Token and Time are reported in tokens and seconds.}
\label{tab:qwen_spider_agent}
\end{table*}

\subsection{Text-to-SQL Performance}
To isolate the downstream effect of schema linking, we keep Spider-Agent and its settings fixed, varying only the schema-linking output provided as input. The results are shown in Table~\ref{tab:qwen_spider_agent}.
On Spider2-Snow, MDB-Link achieves higher EX than both LinkAlign and AutoLink while running faster than either; its token use is much lower than AutoLink's but a bit higher than LinkAlign's. 
On MMQA, MDB-Link improves EX, token use, and runtime over LinkAlign. AutoLink attains higher EX and a shorter runtime but consumes more tokens than MDB-Link, as AutoLink produces a much larger linked schema. 
On BIRD-dev, MDB-Link improves all three downstream measures over LinkAlign; compared with AutoLink, it yields slightly lower EX but uses much fewer tokens (6,007 vs 12,824) and less time with a substantially smaller schema. Complete model-specific results are provided in Appendix~\ref{app:complete_spider_agent_ex}. 
Overall, MDB-Link provides a consistent balance between execution accuracy, schema compactness, and downstream efficiency; it outperforms LinkAlign on accuracy on all three datasets with lower token costs and time in most cases, and often performs competitive as AutoLink on accuracy but costs much fewer tokens and less time.

\section{Ablation Study}
Here, we report and analyse the results of ablation experiments. In Table~\ref{tab:ablation_qwen}, w/o DB reranking removes the LLM-based database reranker and takes the database that has the maximum columns in HRC as the predicted database, whereas w/o table selection removes the intermediate table-selection stage and grounds columns directly over a budget-aware schema context of the predicted database. Complete results for both Qwen2.5-14B and Ministral-3-14B across all three datasets are reported in Appendix~\ref{app:complete_ablation_results}.

\begin{table}[t]
\centering
\small
\setlength{\tabcolsep}{2.2pt}
\begin{tabular*}{\columnwidth}{@{\extracolsep{\fill}}lccc@{}}
\toprule
Method & LA & EM & Recall \\
\midrule
\textbf{MDB-Link} & \textbf{65.00} & \textbf{9.17} & \textbf{32.10} \\
\midrule
w/o DB rerank & 52.50 (-12.50) & 5.83 (-3.34) & 16.78 (-15.32) \\
w/o table selection & 65.00 (0.00) & 7.50 (-1.67) & 21.59 (-10.51) \\
\bottomrule
\end{tabular*}
\caption{Ablation study on Spider2-Snow with Qwen2.5-14B. w/o DB rerank means that the rerank functions are omitted. w/o table selection means that the Table Selection module is omitted.}
\label{tab:ablation_qwen}
\end{table}

Removing database reranking reduces LA by 12.50 points, EM by 3.34 points, and Recall by 15.32 points. This result indicates that retrieval support alone is insufficient to distinguish some semantically similar databases, while LLM reranking provides useful question--schema evidence for the final localization decision. Removing table selection leaves LA unchanged, as expected for a module applied after database localization, but lowers EM by 1.67 points and Recall by 10.51 points. On this large-schema dataset, exposing column grounding to a broader set of tables therefore introduces distractors rather than reliably recovering additional gold columns. The two modules play complementary roles: database reranking determines where to search, while table selection narrows the within-database search space for more reliable column grounding.

\section{Conclusion}
MDB-Link reframes multi-database schema linking as a hierarchical reduction problem rather than flat retrieval over all columns. 
It retrieves question-relevant columns from a global column-wise vector index and aggregates them into database-level evidence, which is further refined through support-based pruning and LLM reranking.
After database localization, table selection, and column-wise grounding then narrow the predicted schema, while budget-aware schema context construction controls the schema context according to the complete prompt budget. 
Together, these components provide a structured alternative to full-schema prompting and unconstrained schema expansion.

Across MMQA, Spider2-Snow, and BIRD-dev, MDB-Link improves LA and downstream EX over LinkAlign in all six datasets: backbone comparisons and improves EM in five of the six; while running faster than LinkAlign and AutoLink in every comparable setting. AutoLink achieves higher recall but selects 9.5--24.9$\times$ more columns than MDB-Link without consistent EX gains.
Ablations support the hierarchical design: removing database reranking lowers LA and Recall in all six settings, while removing table selection lowers EM and Recall in five of six. These results suggest that multi-database schema linking should jointly optimize correctness, compactness, and downstream SQL utility rather than maximize column recall alone.

\section*{Limitations}
In the future, datasets containing larger and more complex schemas are still a challenge for schema linking and text-to-SQL. More advanced methods need to be developed. Also, it needs evaluation and deployment for real-word data environment with larger scale databalse and more complex schemas.

MDB-Link also relies on available schema metadata, including names, descriptions, sample values, and key relations, together with a persistent vector index. In real deployments, such metadata may be incomplete, noisy, or outdated, while changes to database schemas may require the index to be updated. Robustness under these conditions has not yet been systematically evaluated. 
\bibliography{reference}

@inproceedings{lei2025spider,
  title={Spider 2.0: Evaluating language models on real-world enterprise text-to-sql workflows},
  author={Lei, Fangyu and Chen, Jixuan and Ye, Yuxiao and Cao, Ruisheng and Shin, Dongchan and Su, Hongjin and Suo, Zhaoqing and Gao, Hongcheng and Hu, Wenjing and Yin, Pengcheng and others},
  booktitle={International Conference on Learning Representations},
  volume={2025},
  pages={28691--28735},
  year={2025},
  url={https://openreview.net/forum?id=XmProj9cPs}
}

@inproceedings{yu2018spider,
  title={Spider: A large-scale human-labeled dataset for complex and cross-domain semantic parsing and text-to-sql task},
  author={Yu, Tao and Zhang, Rui and Yang, Kai and Yasunaga, Michihiro and Wang, Dongxu and Li, Zifan and Ma, James and Li, Irene and Yao, Qingning and Roman, Shanelle and others},
  booktitle={Proceedings of the 2018 conference on empirical methods in natural language processing},
  pages={3911--3921},
  year={2018}
}

@misc{qwen3technicalreport,
      title={Qwen3 Technical Report}, 
      author={Qwen Team},
      year={2025},
      eprint={2505.09388},
      archivePrefix={arXiv},
      primaryClass={cs.CL},
      url={https://arxiv.org/abs/2505.09388}, 
}

@article{kadavath2022language,
  title={Language Models (Mostly) Know What They Know},
  author={Kadavath, Saurav and Conerly, Tom and Askell, Amanda and Henighan, Tom and Drain, Dawn and Perez, Ethan and Schiefer, Nicholas and Hatfield-Dodds, Zac and DasSarma, Nova and Tran-Johnson, Eli and others},
  journal={arXiv preprint arXiv:2207.05221},
  year={2022},
  url={https://arxiv.org/abs/2207.05221}
}

@article{si2022prompting,
  title={Prompting {GPT}-3 To Be Reliable},
  author={Si, Chenglei and Gan, Zhe and Yang, Zhengyuan and Wang, Shuohang and Wang, Jianfeng and Boyd-Graber, Jordan and Wang, Lijuan},
  journal={arXiv preprint arXiv:2210.09150},
  year={2022},
  url={https://arxiv.org/abs/2210.09150}
}

@article{lewis2020retrieval,
  title={Retrieval-augmented generation for knowledge-intensive nlp tasks},
  author={Lewis, Patrick and Perez, Ethan and Piktus, Aleksandra and Petroni, Fabio and Karpukhin, Vladimir and Goyal, Naman and K{\"u}ttler, Heinrich and Lewis, Mike and Yih, Wen-tau and Rockt{\"a}schel, Tim and others},
  journal={Advances in neural information processing systems},
  volume={33},
  pages={9459--9474},
  year={2020},
  url={https://proceedings.neurips.cc/paper/2020/hash/6b493230-Abstract.html}
}

@inproceedings{karpukhin2020dense,
  title={Dense passage retrieval for open-domain question answering},
  author={Karpukhin, Vladimir and Oguz, Barlas and Min, Sewon and Lewis, Patrick and Wu, Ledell and Edunov, Sergey and Chen, Danqi and Yih, Wen-tau},
  booktitle={Proceedings of the 2020 conference on empirical methods in natural language processing (EMNLP)},
  pages={6769--6781},
  year={2020},
  url={https://aclanthology.org/2020.emnlp-main.550/},
  doi={10.18653/v1/2020.emnlp-main.550}
}

@article{katsogiannis2023survey,
  title={A survey on deep learning approaches for text-to-SQL},
  author={Katsogiannis-Meimarakis, George and Koutrika, Georgia},
  journal={The VLDB Journal},
  volume={32},
  number={4},
  pages={905--936},
  year={2023},
  publisher={Springer},
  doi={10.1007/s00778-022-00776-8},
  url={https://doi.org/10.1007/s00778-022-00776-8}
}

@inproceedings{hui2022s2sql,
    title = "{S}$^2${SQL}: Injecting Syntax to Question-Schema Interaction Graph Encoder for Text-to-{SQL} Parsers",
    author = "Hui, Binyuan  and
      Geng, Ruiying  and
      Wang, Lihan  and
      Qin, Bowen  and
      Li, Yanyang  and
      Li, Bowen  and
      Sun, Jian  and
      Li, Yongbin",
    editor = "Muresan, Smaranda  and
      Nakov, Preslav  and
      Villavicencio, Aline",
    booktitle = "Findings of the Association for Computational Linguistics: ACL 2022",
    month = may,
    year = "2022",
    address = "Dublin, Ireland",
    publisher = "Association for Computational Linguistics",
    url = "https://aclanthology.org/2022.findings-acl.99/",
    doi = "10.18653/v1/2022.findings-acl.99",
    pages = "1254--1262",
}

@inproceedings{cao2021lgesql,
    title = "{LGESQL}: Line Graph Enhanced Text-to-{SQL} Model with Mixed Local and Non-Local Relations",
    author = "Cao, Ruisheng  and
      Chen, Lu  and
      Chen, Zhi  and
      Zhao, Yanbin  and
      Zhu, Su  and
      Yu, Kai",
    editor = "Zong, Chengqing  and
      Xia, Fei  and
      Li, Wenjie  and
      Navigli, Roberto",
    booktitle = "Proceedings of the 59th Annual Meeting of the Association for Computational Linguistics and the 11th International Joint Conference on Natural Language Processing (Volume 1: Long Papers)",
    month = aug,
    year = "2021",
    address = "Online",
    publisher = "Association for Computational Linguistics",
    url = "https://aclanthology.org/2021.acl-long.198/",
    doi = "10.18653/v1/2021.acl-long.198",
    pages = "2541--2555",
}

@inproceedings{wang2020rat,
  title={Rat-sql: Relation-aware schema encoding and linking for text-to-sql parsers},
  author={Wang, Bailin and Shin, Richard and Liu, Xiaodong and Polozov, Oleksandr and Richardson, Matthew},
  booktitle={Proceedings of the 58th annual meeting of the association for computational linguistics},
  pages={7567--7578},
  year={2020},
  url={https://aclanthology.org/2020.acl-main.677/},
  doi={10.18653/v1/2020.acl-main.677}
}

@article{gao2023text,
  title={Text-to-sql empowered by large language models: A benchmark evaluation},
  author={Gao, Dawei and Wang, Haibin and Li, Yaliang and Sun, Xiuyu and Qian, Yichen and Ding, Bolin and Zhou, Jingren},
  journal={arXiv preprint arXiv:2308.15363},
  year={2023},
  url={https://arxiv.org/abs/2308.15363}
}

@inproceedings{sun2023sqlprompt,
  title={Sqlprompt: In-context text-to-sql with minimal labeled data},
  author={Sun, Ruoxi and Arik, Sercan O and Sinha, Rajarishi and Nakhost, Hootan and Dai, Hanjun and Yin, Pengcheng and Pfister, Tomas},
  booktitle={Findings of the Association for Computational Linguistics: EMNLP 2023},
  pages={542--550},
  year={2023},
  url={https://aclanthology.org/2023.findings-emnlp.39/},
  doi={10.18653/v1/2023.findings-emnlp.39}
}

@article{deng2025reforce,
  title={ReFoRCE: a text-to-SQL agent with self-refinement, consensus enforcement, and column exploration},
  author={Deng, Minghang and Ramachandran, Ashwin and Xu, Canwen and Hu, Lanxiang and Yao, Zhewei and Datta, Anupam and Zhang, Hao},
  journal={arXiv preprint arXiv:2502.00675},
  year={2025},
  url={https://arxiv.org/abs/2502.00675}
}

@article{pourreza2023din,
  title={Din-sql: Decomposed in-context learning of text-to-sql with self-correction},
  author={Pourreza, Mohammadreza and Rafiei, Davood},
  journal={Advances in neural information processing systems},
  volume={36},
  pages={36339--36348},
  year={2023},
  url={https://proceedings.neurips.cc/paper_files/paper/2023/hash/72223cc66f63ca1aa59edaec1b3670e6-Abstract-Conference.html}
}

@misc{qwen2.5,
    title = {Qwen2.5: A Party of Foundation Models},
    url = {https://qwenlm.github.io/blog/qwen2.5/},
    author = {Qwen Team},
    month = {September},
    year = {2024}
}

@misc{qdrant2026,
  title        = {Qdrant: Vector Search Engine and Vector Database},
  author       = {{Qdrant}},
  year         = {2026},
  howpublished = {\url{https://qdrant.tech/documentation/}},
  note         = {Accessed: 2026-07-08}
}

@article{liu2026ministral,
  title={Ministral 3},
  author={Liu, Alexander H and Khandelwal, Kartik and Subramanian, Sandeep and Jouault, Victor and Rastogi, Abhinav and Sad{\'e}, Adrien and Jeffares, Alan and Jiang, Albert and Cahill, Alexandre and Gavaudan, Alexandre and others},
  journal={arXiv preprint arXiv:2601.08584},
  year={2026},
  url={https://arxiv.org/abs/2601.08584}
}

@article{qwen3embedding,
  title={Qwen3 Embedding: Advancing Text Embedding and Reranking Through Foundation Models},
  author={Zhang, Yanzhao and Li, Mingxin and Long, Dingkun and Zhang, Xin and Lin, Huan and Yang, Baosong and Xie, Pengjun and Yang, An and Liu, Dayiheng and Lin, Junyang and Huang, Fei and Zhou, Jingren},
  journal={arXiv preprint arXiv:2506.05176},
  year={2025},
  url={https://arxiv.org/abs/2506.05176}
}

@inproceedings{wu2025mmqa,
  title={MMQA: Evaluating LLMs with multi-table multi-hop complex questions},
  author={Wu, Jian and Yang, Linyi and Li, Dongyuan and Ji, Yuliang and Okumura, Manabu and Zhang, Yue},
  booktitle={The thirteenth international conference on learning representations},
  year={2025},
  url={https://openreview.net/forum?id=GGlpykXDCa}
}

@inproceedings{wang2026autolink,
  title={AutoLink: Autonomous Schema Exploration and Expansion for Scalable Schema Linking in Text-to-SQL at Scale},
  author={Wang, Ziyang and Zheng, Yuanlei and Cao, Zhenbiao and Zhang, Xiaojin and Wei, Zhongyu and Fu, Pei and Luo, Zhenbo and Chen, Wei and Bai, Xiang},
  booktitle={Proceedings of the AAAI Conference on Artificial Intelligence},
  volume={40},
  pages={33809--33817},
  year={2026},
  doi={10.1609/aaai.v40i40.40672},
  url={https://ojs.aaai.org/index.php/AAAI/article/view/40672}
}

@inproceedings{wang2025linkalign,
  title={Linkalign: Scalable schema linking for real-world large-scale multi-database text-to-sql},
  author={Wang, Yihan and Liu, Peiyu and Yang, Xin},
  booktitle={Proceedings of the 2025 Conference on Empirical Methods in Natural Language Processing},
  pages={977--991},
  year={2025},
  url={https://aclanthology.org/2025.emnlp-main.51/},
  doi={10.18653/v1/2025.emnlp-main.51}
}

@article{talaei2024chess,
  title={Chess: Contextual harnessing for efficient sql synthesis},
  author={Talaei, Shayan and Pourreza, Mohammadreza and Chang, Yu-Chen and Mirhoseini, Azalia and Saberi, Amin},
  journal={arXiv preprint arXiv:2405.16755},
  year={2024},
  url={https://arxiv.org/abs/2405.16755}
}

@inproceedings{safdarian26schemagraphsql,
    title = "{S}chema{G}raph{SQL}: Efficient Schema Linking with Pathfinding Graph Algorithms for Text-to-{SQL} on Large-Scale Databases",
    author = "Safdarian, AmirHossein  and
      Mohammadi, Milad  and
      Bashirloo, Ehsan Jahanbakhsh  and
      Naderi, Mona Shahamat  and
      Faili, Heshaam",
    editor = "Demberg, Vera  and
      Inui, Kentaro  and
      Marquez, Llu{\'i}s",
    booktitle = "Findings of the {A}ssociation for {C}omputational {L}inguistics: {EACL} 2026",
    month = mar,
    year = "2026",
    address = "Rabat, Morocco",
    publisher = "Association for Computational Linguistics",
    url = "https://aclanthology.org/2026.findings-eacl.134/",
    doi = "10.18653/v1/2026.findings-eacl.134",
    pages = "2585--2599",
    ISBN = "979-8-89176-386-9",
}

@inproceedings{li2023resdsql,
  title={Resdsql: Decoupling schema linking and skeleton parsing for text-to-sql},
  author={Li, Haoyang and Zhang, Jing and Li, Cuiping and Chen, Hong},
  booktitle={Proceedings of the AAAI Conference on Artificial Intelligence},
  volume={37},
  pages={13067--13075},
  year={2023},
  doi={10.1609/aaai.v37i11.26535},
  url={https://ojs.aaai.org/index.php/AAAI/article/view/26535}
}

@inproceedings{li2023bird,
  title     = {Can {LLM} Already Serve as a Database Interface? A {BIg} Bench for Large-Scale Database Grounded Text-to-{SQL}s},
  author    = {Li, Jinyang and Hui, Binyuan and Qu, Ge and Yang, Jiaxi and Li, Binhua and Li, Bowen and Wang, Bailin and Qin, Bowen and Geng, Ruiying and Huo, Nan and Zhou, Xuanhe and Ma, Chenhao and Li, Guoliang and Chang, Kevin and Huang, Fei and Cheng, Reynold and Li, Yongbin},
  booktitle = {Advances in Neural Information Processing Systems},
  volume    = {36},
  year      = {2023},
  url       = {https://proceedings.neurips.cc/paper_files/paper/2023/hash/83fc8fab1710363050bbd1d4b8cc0021-Abstract-Datasets_and_Benchmarks.html}
}

\appendix
\section{Global Column-Wise Vector Index Construction}
\label{app:column_index}

\subsection{Indexed Point and Retrieval Document Examples}
\label{app:embedding_text_example}

Figures~\ref{fig:embedding_text_account_date} and~\ref{fig:embedding_text_card_type} show two indexed column points and their serialized retrieval documents. The examples follow the \texttt{Financial.Account.Date} and \texttt{Financial.Card.Type} columns in Figure~\ref{workflow}. Each vector is computed from the compact retrieval document, while its payload stores provenance and metadata for later database aggregation and schema rendering.

\begin{figure}[ht]
\centering
\begin{minipage}{0.98\linewidth}
\begin{lstlisting}[style=promptbox]
Indexed point:
  column_id: Financial.Account.Date
  payload:
    db_id: Financial
    table_name: Account
    column_name: Date
    data_type: DATE
    sample_values: [2012-08-26,1994-02-08,1995-04-22,
                    1995-09-22,1993-05-09]
    is_primary_key: false
    is_foreign_key: false

Retrieval document:
  Table: Account
  Column: Date
  Description: None
  Value descriptions: in the form YYMMDD
\end{lstlisting}
\end{minipage}
\caption{Indexed point and retrieval document for \texttt{Financial.Account.Date}; \texttt{None} is a literal value in the source metadata and is therefore retained in the serialized retrieval document.}
\label{fig:embedding_text_account_date}
\end{figure}

\begin{figure}[ht]
\centering
\begin{minipage}{0.98\linewidth}
\begin{lstlisting}[style=promptbox]
Indexed point:
  column_id: Financial.Card.Type
  payload:
    db_id: Financial
    table_name: Card
    column_name: Type
    data_type: TEXT
    sample_values: [junior,classic,gold]
    is_primary_key: false
    is_foreign_key: false

Retrieval document:
  Table: Card
  Column: Type
  Description: None
  Value descriptions: "junior": junior class of credit card; "classic": standard class of credit card; "gold": high-level credit card
\end{lstlisting}
\end{minipage}
\caption{Indexed point and retrieval document for \texttt{Financial.Card.Type}.}
\label{fig:embedding_text_card_type}
\end{figure}

\section{Prompt Templates}
\label{app:prompt_templates}

\subsection{LLM-Based Database Reranking Prompt}
\label{app:database_reranking_prompt}

Listing~\ref{lst:database_reranking_prompt} gives the full prompt used for Large Language Model (LLM)-based database reranking. For each candidate database, MDB-Link fills \texttt{\{DATABASE\_SCHEMAS\}} with a budget-aware schema context, \texttt{\{QUESTION\}} with the natural language question, and \texttt{\{HINT\}} with optional external knowledge. The model is constrained to output \texttt{yes} or \texttt{no}; MDB-Link uses the next-token probability of \texttt{yes} as the database relevance score.

\begin{lstlisting}[style=promptbox,caption={Full prompt template for LLM-based database reranking.},label={lst:database_reranking_prompt}]
### Instructions:
You are an expert data analyst specializing in database querying.

Your task is to determine whether the given database is suitable for writing a correct SQL query that answers the question.
You must carefully examine the provided schema context and decide whether it contains the necessary tables, columns, and relationships required to answer the question.
The input schema is a rendered schema context: it begins with the database name, may include primary keys and foreign key relationships, and then lists each table with per-column fields such as column name, data type, description, sample values, and value descriptions.
The context may be filtered or truncated for prompt budget, so use only the tables, columns, and relationships that are explicitly shown.
`NOT_AVAILABLE` means the source data does not provide that metadata.
`NONE` means the source data explicitly indicates there are no items in that section.
Primary keys are listed only for key columns that appear in the schema context.
Foreign key relationships are listed only when both linked columns appear in the schema context.
Return only "yes" or "no".

### Hint:
The hint (if provided) may suggest relevant tables, columns, or relationships.
Use the hint as additional guidance, but rely primarily on the schema context and the question.

### Output Format (STRICT):
- Output must be exactly one word: "yes" or "no"
- Use lowercase only
- Do NOT include any explanations
- Do NOT include reasoning
- Do NOT include JSON
- Do NOT include markdown
- Do NOT include any extra text or symbols

### Decision Rules:
- Answer "yes" if the shown schema context contains sufficient information to construct a correct SQL query for the question.
- Answer "no" if the shown schema context lacks key information required to answer the question.
- Do not assume the existence of hidden tables or columns that are not shown.
- Prefer strict judgment: if unsure, answer "no".
- You may reason internally to determine the answer, but do NOT output the reasoning content.

### INPUT:
Database Schemas:
{DATABASE_SCHEMAS}

Question:
{QUESTION}

Hint (may be empty):
{HINT}

Answer:
\end{lstlisting}

\subsection{Example Candidate-Database Schema Context}
\label{app:database_reranking_example}

Listing~\ref{lst:database_reranking_example} shows a shortened example of the \texttt{\{DATABASE\_SCHEMAS\}} field for the candidate database \texttt{Financial}. Only a small number of tables and columns are shown to illustrate the rendered schema format.

\begin{lstlisting}[style=promptbox,caption={Shortened schema context used as an example input to the database reranking prompt.},label={lst:database_reranking_example}]
Database: Financial

Primary keys:
- Account: Account_id
- District: District_id
- Loan: Loan_id

Foreign key relationships:
- Account.District_id -> District.District_id
- Loan.Account_id -> Account.Account_id

Table: Account
Column: Account_id
Data type: INT
Description: Unique identifier of an account.

Column: District_id
Data type: INT
Description: District associated with the account.

Table: District
Column: A2
Data type: TEXT
Description: District name.
Value descriptions: Contains district information.

Column: A3
Data type: TEXT
Description: Region name.
Value descriptions: Contains region information.

Table: Loan
Column: Loan_id
Data type: INT
Description: Unique identifier of a loan.
\end{lstlisting}

\subsection{Table Selection Prompt}
\label{app:table_selection_prompt}

Listing~\ref{lst:table_selection_prompt} gives the table-selection prompt. The long in-prompt few-shot demonstration schemas are omitted here for readability; the displayed prompt preserves the task instruction, strict output format, and runtime input fields used by MDB-Link.

\begin{lstlisting}[style=promptbox,caption={Prompt template for table selection.},label={lst:table_selection_prompt}]
### Instructions:
You are an expert data analyst specializing in database querying.

Your task is to identify the minimal set of tables needed to write a correct SQL query that answers the question.
The input schema is a rendered schema context: it begins with the database name, may include primary keys and foreign key relationships, and then lists each table with per-column fields such as column name, data type, description, sample values, and value descriptions. The schema context may be filtered or truncated for prompt budget, so only use tables and columns that are explicitly shown. Only include tables that are necessary for the SQL query. Do not include irrelevant tables. Do not rely on tables or columns that are not shown in the schema context.

Important notes:
- The schema context may include only a subset of columns per table because of prompt-budget filtering.
- Primary keys are listed only for key columns that appear in the current schema context.
- Foreign key relationships are listed only when both linked columns appear in the current schema context.
- `Sample values` are separated by semicolons when multiple values are shown.
- `NOT_AVAILABLE` means the source data does not provide that metadata.
- `NONE` means the source data explicitly indicates there are no items in that section.
- Database names, table names, and column names must be interpreted exactly as provided.

### Hint:
The hint (if provided) may mention useful tables, columns, or relationships.
Use it as supplementary guidance, but rely primarily on the schema context and the question.

### Output Requirements (STRICT):
Return ONLY a valid JSON object.

Format:
{
  "relevant_tables": ["table_name1", "table_name2", "table_name3"]
}

Rules (STRICT):
- Each item must be a table name from the provided schema context.
- Include all and only the tables required to write the SQL query.
- Use double quotes for all strings.
- Do NOT invent table names.
- Do NOT include explanations.
- Do NOT include reasoning.
- Do NOT include markdown or code blocks.
- Do NOT include any text before or after the JSON object.

If no table can be confidently identified, return:
{}

You may reason internally to determine the correct tables, but do NOT output the reasoning content.
Only output the final JSON object.

### INPUT:
{DATABASE_SCHEMAS}

Question:
{QUESTION}

Hint (may be empty):
{HINT}
\end{lstlisting}

\subsection{Column-Wise Grounding Prompt}
\label{app:column_grounding_prompt}

Listing~\ref{lst:column_grounding_prompt} gives the column-wise grounding prompt. As above, long few-shot demonstration schemas are omitted to keep the supplementary PDF compact.

\begin{lstlisting}[style=promptbox,caption={Prompt template for column-wise grounding.},label={lst:column_grounding_prompt}]
### Instructions:
You are an expert data analyst specializing in database querying.

Your task is to identify the minimal set of columns used to write a correct SQL query that answers the question.
The database schema provided to you has already been filtered to contain candidate tables for this question. The input schema is a rendered schema context: it begins with the database name, may include primary keys and foreign key relationships, and then lists each table with per-column fields such as column name, data type, description, sample values, and value descriptions. The schema context may be filtered or truncated for prompt budget, so only use tables and columns that are explicitly shown. Do NOT decide which additional tables are relevant. Your job is to identify which visible columns from the provided tables are necessary for the SQL query. Only include columns that are necessary for the SQL query. Do not include irrelevant columns. Do not include extra columns just because they appear in the same table. Do not guess hidden columns that do not appear in the schema context. Include join keys only when they are needed to connect tables in the SQL query and are visible in the schema context.

Important notes:
- The schema context has already been filtered to candidate tables, and it may still include only a subset of columns because of prompt-budget filtering.
- Primary keys are listed only for key columns that appear in the current schema context.
- Foreign key relationships are listed only when both linked columns appear in the current schema context.
- `Sample values` are separated by semicolons when multiple values are shown.
- `NOT_AVAILABLE` means the source data does not provide that metadata.
- `NONE` means the source data explicitly indicates there are no items in that section.
- Table names and column names must be used exactly as provided in the schema context.

### Hint:
The hint (if provided) may highlight columns, conditions, or relationships that are relevant to the question.
Use the hint as additional guidance, but rely primarily on the schema context and the question.

### Output Requirements (STRICT):
Return ONLY a valid JSON object.

Format:
{
  "relevant_columns": {
    "table_name1": ["column1", "column2"],
    "table_name2": ["column3"],
    "table_name3": ["column4", "column5", "column6"]
  }
}

Rules (STRICT):
- Keys must be table names from the provided schema context.
- Values must be lists of column names from the provided schema context.
- Use double quotes for all keys and strings.
- Do NOT invent tables or columns that do not exist in the schema context.
- Do NOT include explanations.
- Do NOT include reasoning.
- Do NOT include markdown or code blocks.
- Do NOT include any text before or after the JSON.
- Only include columns required for writing the SQL query.
- If a required column is not visible in the schema context, do NOT guess it.
- If a table has no required columns, do NOT include that table.

If no columns can be confidently identified, return:
{}

You may reason internally to determine the correct columns, but do NOT output the reasoning content.
Only output the final JSON object.

### INPUT:
{DATABASE_SCHEMAS}

Question:
{QUESTION}

Hint (may be empty):
{HINT}
\end{lstlisting}

\subsection{Single-Prompt Baseline Prompts}
\label{app:single_prompt_prompt}

The Single-Prompt baseline first selects one target database with Listing~\ref{lst:single_prompt_database_prompt}, and then predicts the required columns within that database with Listing~\ref{lst:single_prompt_prompt}. Both templates are reproduced in full. The second-stage schema-linking prompt is the few-shot template used in all Single-Prompt baseline experiments and includes three demonstrations.

\begin{lstlisting}[style=promptbox,caption={Complete database-selection prompt used by the Single-Prompt baseline.},label={lst:single_prompt_database_prompt}]
### Instructions:
You are an expert data analyst specializing in database querying.

Your task is to identify the most relevant database needed to write a correct SQL query that answers the question.

Select only one database that are most likely to contain the necessary tables, columns, and relationships required for answering the question.
Do not select irrelevant databases.

### Database Schemas:
Each database schema describes the structure of one database, including its name, tables, columns, primary keys, foreign keys, and possible join paths.

Table names, column names, and join paths may help indicate whether a database is relevant to the question.
Primary keys uniquely identify rows in a table.
Foreign keys represent relationships between tables and may indicate how tables should be joined.

### Hint:
The hint (if provided) may highlight databases, tables, columns, or relationships that are relevant to the question.
Use the hint as additional guidance, but rely primarily on the database schemas and the question.

Return ONLY a valid JSON object.

Format:
{
    "relevant_database": "database_name"
}

Rules (STRICT):
- The value must be exactly one database name from the provided database schemas.
- Select the single most relevant database for answering the question.
- Use double quotes for all strings.
- Do NOT invent database names that do not exist in the input.
- Do NOT include explanations.
- Do NOT include reasoning.
- Do NOT include markdown or code blocks.
- Do NOT include any text before or after the JSON object.

If no database can be confidently identified, return:
{}

You may reason internally to determine the correct database, but do NOT output the reasoning content.
Only output the final JSON object.

### INPUT:
Database Schemas:
{DATABASE_SCHEMAS}

Question:
{QUESTION}

Hint (may be empty):
{HINT}
\end{lstlisting}

\begin{lstlisting}[style=promptbox,caption={Complete few-shot prompt used in all Single-Prompt baseline experiments.},label={lst:single_prompt_prompt}]
### Instructions:
You are an expert data analyst specializing in database querying.

Your task is to identify the minimal set of columns used to write a correct SQL query that answers the question.
The input schema uses a rendered database format shared across the schema-linking pipeline: it begins with the database name, may include primary keys and foreign key relationships, and then lists each table with per-column fields such as column name, data type, description, sample values, and value descriptions. Use only tables and columns that are explicitly shown in the rendered schema. Only include columns that are necessary for the SQL query. Do not include irrelevant columns. Do not include extra columns just because they appear in the same table. Include join keys only when they are needed to connect tables in the SQL query.

Important notes:
- The schema begins with `Database: ...`, followed by `Primary keys:` and `Foreign key relationships:` sections, and then one or more `Table: ...` blocks.
- Each column is rendered as a block containing `Column`, `Data type`, `Description`, `Sample values`, and `Value descriptions`.
- Primary keys are listed by table in the `Primary keys:` section. If key metadata is unavailable, the section is `NOT_AVAILABLE`. If the source data explicitly indicates there are no primary-key items to show, the section is `NONE`.
- Foreign key relationships are listed as `source_table.source_column -> target_table.target_column`. If relationship metadata is unavailable, the section is `NOT_AVAILABLE`. If the source data explicitly indicates there are no relationships to show, the section is `NONE`.
- `Sample values` are separated by semicolons when multiple values are shown.
- `NOT_AVAILABLE` means the source data does not provide that metadata.
- `NONE` means the source data explicitly indicates there are no items in that section.
- Table names and column names must be used exactly as provided in the rendered schema.

### Hint:
The hint (if provided) may highlight columns, conditions, or relationships that are relevant to the question.
Use the hint as additional guidance, but rely primarily on the rendered schema and the question.

### Output Requirements (STRICT):

Return ONLY a valid JSON object.

Format:
{
  "relevant_columns": {
    "table_name1": ["column1", "column2"],
    "table_name2": ["column3"],
    "table_name3": ["column4", "column5", "column6"]
  }
}

Rules (STRICT):
- Keys must be table names from the provided rendered schema.
- Values must be lists of column names from the rendered schema.
- Use double quotes for all keys and strings.
- Do NOT invent tables or columns that do not exist in the rendered schema.
- Do NOT include explanations.
- Do NOT include reasoning.
- Do NOT include markdown or code blocks.
- Do NOT include any text before or after the JSON.
- Only include columns required for writing the SQL query.
- If a table has no required columns, do NOT include that table.

If no columns can be confidently identified, return:
{}

You may reason internally to determine the correct columns, but do NOT output the reasoning content.
Only output the final JSON object.

### Example 1:
Database: department_management

Primary keys:
- department: Department_ID

Foreign key relationships:
- management.department_ID -> department.Department_ID

Table: department

Column: Department_ID
Data type: INT
Description: Unique identifier for a department.
Sample values: NOT_AVAILABLE
Value descriptions: NOT_AVAILABLE

Column: Name
Data type: TEXT
Description: Department name.
Sample values: NOT_AVAILABLE
Value descriptions: NOT_AVAILABLE

Column: Budget_in_Billions
Data type: REAL
Description: Department budget.
Sample values: NOT_AVAILABLE
Value descriptions: NOT_AVAILABLE

Column: Num_Employees
Data type: REAL
Description: Number of employees.
Sample values: NOT_AVAILABLE
Value descriptions: NOT_AVAILABLE

Table: management

Column: department_ID
Data type: INT
Description: Department managed by a head.
Sample values: NOT_AVAILABLE
Value descriptions: NOT_AVAILABLE

Column: head_ID
Data type: INT
Description: Head managing the department.
Sample values: NOT_AVAILABLE
Value descriptions: NOT_AVAILABLE

Column: temporary_acting
Data type: TEXT
Description: Whether the head is temporary acting.
Sample values: NOT_AVAILABLE
Value descriptions: NOT_AVAILABLE

Question:
Which department currently headed by a temporary acting manager has the largest number of employees, and how many employees does it have?

Hint (may be empty):
No hint

response:
{
  "relevant_columns": {
    "department": ["Department_ID", "Name", "Num_Employees"],
    "management": ["department_ID", "temporary_acting"]
  }
}

### Example 2:
Database: activity_1

Primary keys:
- Student: StuID
- Activity: actid

Foreign key relationships:
- Participates_in.stuid -> Student.StuID
- Participates_in.actid -> Activity.actid

Table: Student

Column: StuID
Data type: integer
Description: Unique identifier for a student.
Sample values: 1001; 1002; 1003
Value descriptions: NOT_AVAILABLE

Column: LName
Data type: text
Description: Last name of the student.
Sample values: Smith; Kim; Jones
Value descriptions: NOT_AVAILABLE

Column: Fname
Data type: text
Description: First name of the student.
Sample values: Linda; Tracy; Shiela
Value descriptions: NOT_AVAILABLE

Column: Age
Data type: integer
Description: Age of the student.
Sample values: 18; 19; 21
Value descriptions: NOT_AVAILABLE

Table: Participates_in

Column: stuid
Data type: integer
Description: Student participating in an activity.
Sample values: 1001; 1002
Value descriptions: NOT_AVAILABLE

Column: actid
Data type: integer
Description: Activity joined by the student.
Sample values: 770; 771; 777
Value descriptions: NOT_AVAILABLE

Table: Activity

Column: actid
Data type: integer
Description: Unique identifier for an activity.
Sample values: 770; 771; 772
Value descriptions: NOT_AVAILABLE

Column: activity_name
Data type: text
Description: Name of the activity.
Sample values: Mountain Climbing; Canoeing; Kayaking
Value descriptions: NOT_AVAILABLE

Question:
Which activities does student Linda Smith participate in?

Hint (may be empty):
No hint

response:
{
  "relevant_columns": {
    "Student": ["StuID", "LName", "Fname"],
    "Participates_in": ["stuid", "actid"],
    "Activity": ["actid", "activity_name"]
  }
}

### Example 3:
Database: ADVENTUREWORKS

Primary keys:
NOT_AVAILABLE

Foreign key relationships:
NOT_AVAILABLE

Table: ADVENTUREWORKS.ADVENTUREWORKS.CURRENCYRATE

Column: currencyrateid
Data type: INT
Description: NOT_AVAILABLE
Sample values: 2; 4; 5
Value descriptions: NOT_AVAILABLE

Column: tocurrencycode
Data type: STRING
Description: NOT_AVAILABLE
Sample values: AUD; CAD; CNY
Value descriptions: NOT_AVAILABLE

Column: endofdayrate
Data type: FLOAT
Description: NOT_AVAILABLE
Sample values: 1.55; 1.4683; 8.2784
Value descriptions: NOT_AVAILABLE

Column: fromcurrencycode
Data type: STRING
Description: NOT_AVAILABLE
Sample values: USD
Value descriptions: NOT_AVAILABLE

Column: averagerate
Data type: FLOAT
Description: NOT_AVAILABLE
Sample values: 1.5491; 1.4641; 8.2781
Value descriptions: NOT_AVAILABLE

Question:
What are the average and end-of-day exchange rates from USD to CAD?

Hint (may be empty):
No hint

response:
{
  "relevant_columns": {
    "ADVENTUREWORKS.ADVENTUREWORKS.CURRENCYRATE": ["fromcurrencycode", "tocurrencycode", "averagerate", "endofdayrate"]
  }
}

### INPUT:
{DATABASE_SCHEMAS}

Question:
{QUESTION}

Hint (may be empty):
{HINT}
\end{lstlisting}

\section{Budget-Aware Schema Context Construction}
\label{app:budget_aware_context}

\subsection{Construction Procedure}
\label{app:budget_aware_context_procedure}

MDB-Link uses two related procedures to construct the schema context that fills \texttt{\{DATABASE\_SCHEMAS\}}. Database reranking and table selection use the shared full-schema-first renderer over one candidate or predicted database (Algorithm~\ref{alg:budget_aware_context}). Column-wise grounding uses the selected-table context procedure: it first tests all columns of the selected tables and otherwise reuses the table-selection records retained for those tables when available (Algorithm~\ref{alg:column_context_fallback}). In both procedures, the token test is performed on the complete prompt, including the prompt template, question, optional hint, and schema context.

\begin{algorithm}[H]
\small
\caption{Full-Schema-First Renderer}
\label{alg:budget_aware_context}
\begin{algorithmic}[1]
\REQUIRE Database \(D\), question \(q\), optional hint \(h\), prompt template \(P\), prompt cap \(B\), column index \(\mathcal I\)
\ENSURE Rendered schema context \(C\) for \texttt{\{DATABASE\_SCHEMAS\}}
\STATE Render the complete database schema \(S_{\text{full}}\), preserving source order and available metadata.
\IF{\(\mathrm{tokens}(P(S_{\text{full}},q,h)) \leq B\)}
    \RETURN \(S_{\text{full}}\)
\ENDIF
\STATE Embed \(q\) and retrieve all indexed columns of \(D\); assign each column its cosine-similarity score.
\STATE Initialize \(A\) with the highest-scoring column of every table; mark these columns as protected.
\STATE Add all available primary-key and foreign-key columns to \(A\).
\IF{\(\mathrm{tokens}(P(\mathrm{Render}(A),q,h))>B\)}
    \STATE Remove unprotected foreign keys from \(A\) in ascending score order until the prompt fits or none remain.
    \STATE Then remove unprotected primary keys in ascending score order under the same condition.
\ENDIF
\IF{the prompt still exceeds \(B\)}
    \RETURN \(\mathrm{Render}(A)\)
\ENDIF
\STATE Within each table, sort unselected columns by decreasing score (breaking ties by source order).
\LOOP
    \STATE Form a round \(U\) from the next unselected column of every table.
    \IF{\(U=\emptyset\) or \(\mathrm{tokens}(P(\mathrm{Render}(A\cup U),q,h))>B\)}
        \STATE \textbf{break}
    \ENDIF
    \STATE \(A\leftarrow A\cup U\)
\ENDLOOP
\STATE Render \(C\) from \(A\) in source schema order; show a foreign-key relation only when both endpoints are retained.
\RETURN \(C\)
\end{algorithmic}
\end{algorithm}

For database reranking, hint handling is applied around the shared renderer. If the prompt without schema already exceeds \(B\) when \(h\) is included, MDB-Link replaces the hint with \texttt{No hint}. After schema construction, it applies the same replacement if the complete prompt still exceeds \(B\). A candidate database whose prompt remains over the cap receives no reranking score and is placed after scored candidates.

\begin{algorithm}[H]
\small
\caption{Selected-Table Context Procedure for Column Grounding}
\label{alg:column_context_fallback}
\begin{algorithmic}[1]
\REQUIRE Predicted database \(\hat D\), selected tables \(\hat{\mathcal T}\), table-stage retained records \(A_T\), question \(q\), hint \(h\), column prompt \(P_C\), prompt cap \(B\)
\ENSURE Column-grounding schema context \(C_C\)
\STATE \textbf{if} \(\hat{\mathcal T}=\emptyset\) \textbf{then return} \(\mathrm{Render}(A_T)\)
\STATE Let \(A_{\mathrm{full}}\) contain every column of tables in \(\hat{\mathcal T}\).
\STATE \textbf{if} \(A_{\mathrm{full}}\neq\emptyset\) and \(\mathrm{tokens}(P_C(\mathrm{Render}(A_{\mathrm{full}}),q,h))\leq B\) \textbf{then return} \(\mathrm{Render}(A_{\mathrm{full}})\)
\STATE \(A_C\leftarrow\{c\in A_T:\mathrm{table}(c)\in\hat{\mathcal T}\}\)
\STATE \textbf{if} \(A_C=\emptyset\) \textbf{then} \(A_C\leftarrow A_T\)
\STATE \textbf{if} \(A_C=\emptyset\) \textbf{then} \(A_C\leftarrow A_{\mathrm{full}}\)
\RETURN \(\mathrm{Render}(A_C)\)
\end{algorithmic}
\end{algorithm}

\section{Fixed Implementation Settings}
\label{app:implementation_settings}

For database localization, both retrieval rounds use the HRC ratio \(\alpha=0.1\). The first- and second-round retrieval caps are \(\beta_1=500\) and \(\beta_2=50\), respectively. \(\mathrm{PruneDB}\) uses the minimum hit count \(\eta=2\), the similarity quantile \(\rho=0.8\), and the cap \(\mu=10\). The second round is triggered when more than \(\kappa=3\) databases remain, in which case the reranker retains the top-\(\kappa\) databases as its search scope.

Table~\ref{tab:implementation_settings} reports the remaining fixed hyperparameter and inference settings used in all experiments.

\begin{table*}[t]
\centering
\small
\setlength{\tabcolsep}{6pt}
\begin{tabular}{p{0.22\textwidth}p{0.72\textwidth}}
\toprule
Component & Fixed settings \\
\midrule
Prompting & Database reranking uses the zero-shot binary prompt; table selection and column-wise grounding use the few-shot templates. \\
Schema context construction & Maximum LLM input length \(L=110{,}000\) tokens; complete-prompt cap \(\min(\lfloor0.85L\rfloor,L-512)\). \\
Spider-Agent & The Agent receives only the schema items returned by the linker. It uses a local Transformers backend with bfloat16 precision with an 80,000-token input limit, a 4,096-token output limit, a 12,000-token history budget, at most 20 rounds, two retries after a failed model call, one rollout per example, temperature 0, and random seed 42. \\
Model use & All models are used for inference without parameter updates. Applicable settings are fixed across datasets, and the Spider-Agent settings are fixed across linkers. \\
\bottomrule
\end{tabular}
\caption{Fixed hyperparameter and inference settings.}
\label{tab:implementation_settings}
\end{table*}

\section{Detailed Dataset Statistics}
\label{app:detailed_dataset_statistics}

Table~\ref{tab:dataset_difficulty} reports complementary statistics on schema scale and the gold schema required per example.

\begin{table*}[!t]
\centering
\small
\begin{tabular}{lrrrr}
\toprule
Dataset & Tables\#/Cols\# per DB & Cols\# per Table & Gold Tables\# per Sample & Gold Cols\# per Sample \\
\midrule
MMQA & 5.11/26.71 & 5.22 & 2.21 & 5.28 \\
Spider2-Snow & 36.57/1,930.96 & 52.81 & 2.63 & 9.88 \\
BIRD (dev) & 6.82/72.55 & 10.64 & 1.94 & 4.47 \\
\bottomrule
\end{tabular}
\caption{Detailed statistics of the three evaluation datasets. Cols\# denotes the number of columns; BIRD refers to the development split.}
\label{tab:dataset_difficulty}
\end{table*}

\section{Complete Efficiency Results}
\label{app:complete_efficiency_results}

Table~\ref{tab:complete_efficiency_results} reports both 14B backbones. Token usage is compared only within the same dataset and backbone and is interpreted jointly with online runtime and schema-linking quality. MDB-Link is faster than LinkAlign and AutoLink in all six matched comparisons. Compared with LinkAlign, it improves LA in all six comparisons, EM in five, and Recall in four; its token usage is lower in four comparisons but higher on Spider2-Snow under both backbones. Compared with AutoLink, MDB-Link uses fewer tokens in four comparisons and more on BIRD-dev under both backbones, while AutoLink attains higher Recall with substantially broader schemas. Single-Prompt has the lowest runtime where feasible, but MDB-Link yields stronger schema-linking quality and uses fewer tokens on MMQA under both backbones; Single-Prompt cannot process Spider2-Snow. Overall, MDB-Link provides a favorable balance between online efficiency and schema-linking quality rather than uniformly minimizing token usage or runtime.

\begin{table*}[!t]
\centering
\small
\setlength{\tabcolsep}{4.5pt}
\begin{tabular*}{\textwidth}{@{\extracolsep{\fill}}lrrrrrr@{}}
\toprule
& \multicolumn{2}{c}{Spider2-Snow} & \multicolumn{2}{c}{MMQA} & \multicolumn{2}{c}{BIRD-dev} \\
\cmidrule(lr){2-3}\cmidrule(lr){4-5}\cmidrule(lr){6-7}
Method & Tok. & Time & Tok. & Time & Tok. & Time \\
\midrule
LinkAlign (M3) & 476 & 743.28 & 122 & 169.25 & 145 & 176.43 \\
LinkAlign (Q2.5) & 223 & 409.72 & 118 & 119.35 & 122 & 127.62 \\
\addlinespace[1pt]
AutoLink (M3) & 635 & 637.68 & 83 & 105.49 & 38 & 131.03 \\
AutoLink (Q2.5) & 652 & 465.42 & 118 & 64.84 & 48 & 45.22 \\
\addlinespace[1pt]
Single-Prompt (M3) & -- & -- & 48 & 11.37 & 14 & 3.79 \\
Single-Prompt (Q2.5) & -- & -- & 45 & 12.28 & 13 & 3.82 \\
\midrule
\textbf{MDB-Link (M3)} & 495 & 291.64 & 41 & 18.33 & 58 & 13.74 \\
\textbf{MDB-Link (Q2.5)} & 515 & 369.17 & 40 & 14.15 & 58 & 12.30 \\
\midrule
Dense Retrieval & 0 & 1.18 & 0 & 1.16 & -- & 0.03 \\
\bottomrule
\end{tabular*}
\caption{Complete efficiency results under Ministral-3-14B (M3) and Qwen2.5-14B (Q2.5). Tokens are the average LLM token count per example, reported in thousands, and Time is the average runtime per example in seconds. Token values are comparable only within the same dataset and backbone; a dash denotes unavailable or infeasible results. Dense Retrieval is retrieval-only.}
\label{tab:complete_efficiency_results}
\end{table*}

\section{Complete Spider-Agent Results}
\label{app:complete_spider_agent_ex}

Table~\ref{tab:complete_spider_agent} reports the complete downstream results for the Ministral-3-14B and Qwen2.5-14B schema-linking backbones. Spider-Agent and its configuration are fixed across methods, so the only changing input is the schema returned by each linker.

Across the two backbones, MDB-Link improves EX over LinkAlign in all six dataset--backbone comparisons. On MMQA and BIRD-dev, it also uses fewer SQL-generation tokens and less time under both backbones. The efficiency pattern on Spider2-Snow is backbone-dependent: MDB-Link is faster but uses more tokens with Qwen2.5-14B, whereas it uses more tokens and time with Ministral-3-14B. The EX gains therefore remain consistent even when the downstream cost advantage does not.

Compared with AutoLink, MDB-Link obtains higher EX on Spider2-Snow under both backbones while using 43.8--46.2\% fewer tokens and less time. AutoLink instead achieves higher EX on MMQA and BIRD-dev, but MDB-Link uses 26.6--52.6\% fewer tokens and is faster in three of these four comparisons. Thus, AutoLink's broader schemas can recover useful evidence in some cases, but consistently increase token demand. Overall, the results support evaluating schema linkers jointly by execution accuracy, token use, and runtime rather than treating linked-schema size as a sufficient predictor of downstream behavior.

\begin{table*}[!t]
\centering
\small
\setlength{\tabcolsep}{3.0pt}
\begin{tabular*}{\textwidth}{@{\extracolsep{\fill}}llrrrrrrrrr@{}}
\toprule
& & \multicolumn{3}{c}{Spider2-Snow} & \multicolumn{3}{c}{MMQA} & \multicolumn{3}{c}{BIRD-dev} \\
\cmidrule(lr){3-5}\cmidrule(lr){6-8}\cmidrule(lr){9-11}
Model & Method & EX & Tok. & Time & EX & Tok. & Time & EX & Tok. & Time \\
\midrule
M3 & LinkAlign & 3.33 & \textbf{50,959} & \textbf{160.14} & 60.07 & 6,128 & 24.53 & 55.74 & 5,995 & 23.27 \\
M3 & AutoLink & 4.17 & 100,365 & 175.72 & \textbf{69.57} & 7,127 & 22.27 & \textbf{57.30} & 11,660 & 23.21 \\
\cmidrule(lr){2-11}
M3 & \textbf{MDB-Link} & \textbf{6.67} & 56,429 & 168.60 & 61.45 & \textbf{5,231} & \textbf{21.11} & 55.80 & \textbf{5,920} & \textbf{22.45} \\
\midrule
Q2.5 & LinkAlign & 7.50 & \textbf{46,235} & 181.56 & 55.40 & 7,169 & 30.58 & 50.00 & 7,248 & 26.47 \\
Q2.5 & AutoLink & 7.50 & 103,307 & 163.50 & \textbf{69.18} & 7,662 & \textbf{22.79} & \textbf{57.69} & 12,824 & 25.71 \\
\cmidrule(lr){2-11}
Q2.5 & \textbf{MDB-Link} & \textbf{8.30} & 55,532 & \textbf{144.58} & 63.04 & \textbf{5,567} & 25.71 & 55.35 & \textbf{6,077} & \textbf{21.53} \\
\bottomrule
\end{tabular*}
\caption{Complete Spider-Agent results under Ministral-3-14B (M3) and Qwen2.5-14B (Q2.5) schema-linking backbones. EX is a percentage; Tokens and Time are the average Spider-Agent token count and runtime per example, with runtime reported in seconds. Bold indicates the best result within each backbone and column.}
\label{tab:complete_spider_agent}
\end{table*}

\section{Complete Ablation Results}
\label{app:complete_ablation_results}

Table~\ref{tab:complete_ablation_results} reports the complete ablation results for Ministral-3-14B and Qwen2.5-14B. The two ablations follow the definitions in the main paper: w/o DB reranking replaces the LLM reranker with the support-based pruning order, and w/o table selection grounds columns directly over the predicted database without first restricting the table scope.

\begin{table*}[!t]
\centering
\small
\setlength{\tabcolsep}{4.5pt}
\begin{tabular*}{\textwidth}{@{\extracolsep{\fill}}lllrrr@{}}
\toprule
Model & Dataset & Variant & LA & EM & Recall \\
\midrule
M3 & Spider2-Snow & \textbf{MDB-Link} & \textbf{60.00} & \textbf{1.67} & \textbf{39.12} \\
M3 & Spider2-Snow & w/o DB reranking & 52.50 (-7.50) & 3.33 (+1.66) & 18.80 (-20.32) \\
M3 & Spider2-Snow & w/o table selection & 60.00 (0.00) & 1.67 (0.00) & 21.25 (-17.87) \\
\addlinespace[1pt]
M3 & MMQA & \textbf{MDB-Link} & \textbf{90.54} & \textbf{43.00} & \textbf{83.22} \\
M3 & MMQA & w/o DB reranking & 76.06 (-14.48) & 36.54 (-6.46) & 69.15 (-14.07) \\
M3 & MMQA & w/o table selection & 90.54 (0.00) & 35.93 (-7.07) & 81.79 (-1.43) \\
\addlinespace[1pt]
M3 & BIRD-dev & \textbf{MDB-Link} & \textbf{98.89} & \textbf{39.05} & \textbf{82.72} \\
M3 & BIRD-dev & w/o DB reranking & 96.48 (-2.41) & 36.07 (-2.98) & 79.19 (-3.53) \\
M3 & BIRD-dev & w/o table selection & 98.89 (0.00) & 35.14 (-3.91) & 86.29 (+3.57) \\
\midrule
Q2.5 & Spider2-Snow & \textbf{MDB-Link} & \textbf{65.00} & \textbf{9.17} & \textbf{32.10} \\
Q2.5 & Spider2-Snow & w/o DB reranking & 52.50 (-12.50) & 5.83 (-3.34) & 16.78 (-15.32) \\
Q2.5 & Spider2-Snow & w/o table selection & 65.00 (0.00) & 7.50 (-1.67) & 21.59 (-10.51) \\
\addlinespace[1pt]
Q2.5 & MMQA & \textbf{MDB-Link} & \textbf{89.71} & \textbf{51.41} & \textbf{86.32} \\
Q2.5 & MMQA & w/o DB reranking & 75.29 (-14.42) & 42.84 (-8.57) & 71.11 (-15.21) \\
Q2.5 & MMQA & w/o table selection & 89.71 (0.00) & 42.14 (-9.27) & 84.05 (-2.27) \\
\addlinespace[1pt]
Q2.5 & BIRD-dev & \textbf{MDB-Link} & \textbf{98.89} & \textbf{38.01} & \textbf{81.19} \\
Q2.5 & BIRD-dev & w/o DB reranking & 95.20 (-3.69) & 36.83 (-1.18) & 78.77 (-2.42) \\
Q2.5 & BIRD-dev & w/o table selection & 98.89 (0.00) & 36.25 (-1.76) & 80.20 (-0.99) \\
\bottomrule
\end{tabular*}
\caption{Complete ablation results under Ministral-3-14B (M3) and Qwen2.5-14B (Q2.5). LA, EM, and Recall are percentages; parentheses report changes in percentage points from each bold MDB-Link reference row.}
\label{tab:complete_ablation_results}
\end{table*}

Removing database reranking lowers LA and Recall in all six dataset--backbone comparisons, with LA decreases of 2.41--14.48 points and Recall decreases of 2.42--20.32 points. EM also decreases in five comparisons; the exception is Spider2-Snow with Ministral-3-14B, where EM increases from 1.67 to 3.33 despite substantial reductions in LA and Recall. These results consistently associate the reranker with more reliable database localization, while also showing that no single column-set metric fully characterizes localization quality.

Removing table selection does not change LA because it operates after database localization. It reduces EM in five comparisons and leaves it unchanged in one, while Recall decreases in five of the six comparisons. The effect is largest on Spider2-Snow, where Recall falls by 17.87 points with Ministral-3-14B and 10.51 points with Qwen2.5-14B. The exception is BIRD-dev with Ministral-3-14B, where the broader table scope raises Recall but lowers EM. Overall, table selection generally makes column grounding more reliable by removing distractor tables, although the BIRD-dev result also reflects the expected trade-off between broader coverage and exact schema selection.

\end{document}